# MACHINE LEARNING MODELING FOR REAL-TIME MELT POOL MONITORING IN LASER POWDER BED FUSION ADDITIVE MANUFACTURING: A HYBRID APPROACH

**Inioluwa Emmanuel**
Industrial and
Manufacturing Engineering
Florida State University
Tallahassee, FL

**Zhuo Yang**
National Institute of
Standards and Technology
Gaithersburg, Maryland

**Ho Yeung**
National Institute of
Standards and Technology
Gaithersburg, Maryland

**Xinyao Zhang***
Industrial and
Manufacturing Engineering
Florida State University
Tallahassee, FL
xz25f@fsu.edu

## ABSTRACT

*This work investigates the implementation of artificial intelligence and machine learning (AI/ML) for real-time monitoring in laser powder bed fusion (LPBF) additive manufacturing. We developed a binary image classification framework for distinguishing normal and abnormal melt pool images using a balanced dataset of 1,200 images collected from Nickel superalloy 625 on the NIST AMMT platform. The study evaluates accuracy and inference time based on control requirements and hardware limitations of open-architecture LPBF machines. We benchmark three transfer learning architectures (ResNet50, EfficientNetB0, and MobileNetV2) against two Random Forest approaches: one trained on EfficientNetB0 feature embeddings (hybrid) and one trained on raw pixel features (baseline). Images are stratified into 80/20 train–test splits, with a further 90/10 validation split on the training set, and undergo standardized resizing, normalization, and label-preserving data augmentation to emulate realistic process variability. Each model is evaluated using accuracy, precision, recall, F1 score, and area under the receiver operating characteristic curve (AUC), along with training time, inference latency, and CPU & GPU usage to capture deployability constraints relevant to factory-floor monitoring. The hybrid EfficientNetB0-plus-Random Forest approach achieves the best performance on the held-out test set, with an F1 score of 0.9451, accuracy of 0.9458, and AUC of 0.9904, while maintaining sub-millisecond per-image inference (1.15 ms). In contrast, purely deep learning models exhibit significantly higher inference times with lower accuracy. These results demonstrate that combining pretrained convolutional features with classical ensemble methods provides a robust, computationally efficient route to real-time melt pool anomaly detection in data-limited additive manufacturing environments.*

**Keywords**: Additive manufacturing, Laser powder bed fusion, Melt pool monitoring, Anomaly detection, Hybrid deep learning.

## 1. INTRODUCTION

Laser powder bed fusion (LPBF) has emerged as a transformative additive manufacturing (AM) technology enabling the production of complex metal components with unprecedented geometric freedom [1]. However, the technology faces persistent quality challenges: melt pool anomalies such as keyholing, lack of fusion, and unstable geometries serve as precursors to porosity and crack formation, compromising structural integrity and functional performance [21,24]. Understanding and controlling melt pool dynamics is therefore critical for process optimization and reliable part production [1,3,4], with continuous monitoring of melt pool morphology identified as essential for advancing LPBF industrial deployment [25].

Machine learning (ML) has demonstrated significant potential for real-time quality assurance in LPBF [4,5,10], with deep learning methods achieving high accuracy in melt pool anomaly detection [6,7,8,11]. However, a critical gap exists between research demonstrations and industrial deployment: purely deep learning models often require computational resources and inference times incompatible with millisecond-level process control requirements. Conversely, classical ML approaches offer computational efficiency but lack the representational capacity to extract meaningful features directly from raw image data [16,18]. This accuracy-speed trade-off, combined with limited systematic comparisons across architectures, hinders informed model selection for real-time LPBF monitoring.

Hybrid architectures combining pre-trained convolutional neural network (CNN) feature extractors with classical ML classifiers have proven effective in medical imaging and remote sensing [2,16,18,20,22], yet their application to LPBF melt pool classification remains unexplored. The fundamental question for industrial deployment is whether these hybrid approaches can achieve the representational power of deep learning while maintaining the computational efficiency required for real-time manufacturing control, and how they compare quantitatively to purely deep learning alternatives under identical experimental conditions.

This study provides a rigorous comparative evaluation of ML architectures for LPBF melt pool classification, explicitly addressing the accuracy-speed trade-off critical for industrial deployment. We benchmark three transfer learning models (ResNet50, EfficientNetB0, MobileNetV2) against a hybrid EfficientNetB0-Random Forest pipeline and a raw pixel baseline under controlled experimental conditions with a balanced 1,200-image dataset. Our systematic evaluation quantifies both classification performance (accuracy, precision, recall, F1, AUC) and training time, inference latency, and resource usage. Results demonstrate that the hybrid approach achieves test performance (F1 = 0.9451, AUC = 0.9904) with 1.15 ms inference time, 74 times faster than EfficientNetB0 transfer learning (85.47 ms) and 156 times faster than ResNet50 (179.51 ms), while improving accuracy, establishing empirical evidence for hybrid architectures as the preferred solution for data-limited, real-time LPBF quality monitoring.

While this study offers empirical support for hybrid architectures in real-time LPBF melt pool monitoring, the findings are considered in along with key limitations, including the modest dataset size, single-material with single-machine scope, and static analysis. These are discussed in detail in the Summary section.

## 2. LITERATURE REVIEW

### 2.1 Machine-learning-powered Melt Pool Monitoring and Anomaly Detection in LPBF

ML-based monitoring of laser powder bed fusion has enabled real-time assessment of process quality through analysis of sensor data [10], with comprehensive reviews identifying numerous opportunities for integrating data-driven approaches into LPBF process monitoring and optimization [5]. In-situ melt pool measurements using multi-sensing and correlation analysis have demonstrated the feasibility of capturing critical process signatures [4], while methodologies for determining melt pool anomalies through sensor data fusion have advanced real-time quality assessment capabilities [21,25]. Multi-modal sensing approaches, including correlation between in-situ pyrometry and ex-situ X-ray radiography, have validated that observable melt pool characteristics are predictive of defect formation [23,24].

### 2.2 Deep Learning Approaches for Melt Pool Image Classification

Convolutional neural networks have emerged as the dominant architecture for automated melt pool image classification in LPBF. Early investigations demonstrated the feasibility of using CNNs to distinguish between normal and abnormal melt pool states with minimal feature engineering [7], and subsequent work was able to achieve robust performance across different operating conditions [6]. Engineering-guided deep learning approaches have incorporated domain knowledge about melt pool dynamics into model architectures, improving both accuracy and interpretability [8]. Transfer learning with pre-trained CNNs has proven particularly effective for quality inspection tasks where labeled training data may be limited [11], while data-driven methods have extended to forecasting melt pool characteristics based on previously built parts [19].

Specific to LPBF anomaly detection, combined unsupervised and supervised ML techniques have been applied to melt pool images [14], and data-efficient sequential learning frameworks have been proposed to address limited labeled data challenges [15]. In-situ thermal imaging methods for defect detection have demonstrated transferability across different AM technologies [12]. Despite these successes, a critical limitation of purely deep learning approaches is the trade-off between model accuracy and inference speed, which becomes particularly important for real-time industrial deployment where millisecond-level decision-making is required.

### 2.3 Hybrid Deep Learning in Image Classification

The integration of deep learning feature extraction with classical ML classifiers represents an emerging paradigm for achieving high accuracy while maintaining low inference times. In medical imaging, hybrid architectures combining pre-trained CNN feature extractors with random forest classifiers have demonstrated superior performance for diabetic retinopathy detection [16], pneumonia classification [20], nonalcoholic fatty liver disease grading [2], and skin cancer classification [18]. These hybrid approaches exploit the hierarchical feature learning of deep networks while leveraging the sample efficiency and fast inference of tree-based ensemble methods. Ensemble methods incorporating multiple CNNs have also proven effective, providing robustness through model diversity [el17].

The effectiveness of hybrid techniques has been demonstrated beyond medical imaging, with comparative studies in remote sensing showing that combining deep feature extraction with classical ML can yield superior performance compared to either approach alone [22]. The key advantage of this hybrid paradigm is the decoupling of representation learning from classification: the CNN learns rich, transferable features from images, while the classical ML classifier makes rapid predictions on these features without the computational overhead of deep network inference. This architecture is particularly attractive for applications requiring the best blend of high accuracy with low inference time, as the feature extraction can be optimized for accuracy while the classifier is optimized for speed. For LPBF melt pool monitoring, this approach offers the potential to achieve the classification performance of state-of-the-art deep learning models while meeting the strict latency requirements of real-time process control.

### 2.4 Research Gaps and Study Motivation

Despite progress in applying ML to LPBF quality monitoring, several limitations motivate the present study. First, while deep learning models have achieved high accuracy for melt pool anomaly detection [6,7,8], their computational requirements and inference latency are rarely measured or reported. Most studies describe real-time monitoring as a motivation, yet few provide millisecond-level timing benchmarks, CPU/GPU usage, or deployability analysis. Real-time process control in LPBF requires decisions at millisecond timescales, making the absence of systematic latency evaluation a barrier to practical adoption.

Second, most existing work has focused exclusively on either deep learning or classical approaches, with limited exploration of hybrid architectures combining both paradigms. Hybrid pipelines, where pretrained CNNs extract features and classical models such as Random Forests perform classification, have demonstrated strong performance in medical imaging and remote sensing [2,16,18,20,22]. However, their application to LPBF melt pool classification remains largely unexplored, despite their potential to combine deep learning's representational power with the speed and sample efficiency of classical ML.

Third, many LPBF studies evaluate a single model or compare methods under inconsistent experimental conditions [9,14,15], making cross-study comparisons unreliable. There is a shortage of studies that benchmark multiple architectures, deep learning, classical ML, and hybrid models, under identical preprocessing, dataset splits, and evaluation metrics. Without standardized comparisons, identifying architectures that best balance accuracy, robustness, and deployability remains difficult.

This study addresses these gaps by conducting a rigorous comparative evaluation of deep learning and hybrid machine-learning approaches for LPBF melt pool classification, with explicit consideration of both accuracy and inference speed. By benchmarking transfer-learning models (ResNet50, EfficientNetB0, MobileNetV2) alongside a novel hybrid EfficientNetB0-Random Forest architecture under controlled experimental conditions, we provide empirical evidence for selecting models that achieve the optimal blend of high accuracy and low inference time for real-time quality monitoring in metal AM.

## 3. METHODOLOGY

### 3.1 Dataset

The experiment is conducted on the AM Metrology Testbed (AMMT) at National Institute of Standards and Technology (NIST). The AMMT [27] is a fully customized metrology instrument that enables flexible control and measurement of the L-PBF process. An in-house developed AM software (SAM), which is capable of stereolithography (STL) slicing, scan path planning, and G code generation and interpretation [28], was used to program the different scan strategies for the experiment. Nickel superalloy 625 powder and substrate were used. Twelve rectangular parts (5 mm x 5 mm x 9 mm), with chamfered corners, were laid on the substrate, with a minimum spacing of 10 mm between parts. Corners of the rectangular parts were chamfered by 1 mm at 45°. Each part was built with a different scan strategy. The melt pool was monitored by a high-speed camera which is optically aligned with the heating laser, such that the image of the melt pool is maintained stationary within the camera's field of view. The camera can be triggered up to 100,000 images per second, with an integration time of 20 $\mu s$. The experimental platform used in this study, including the imaging system and control software, is based on the architecture and calibration procedures detailed in [29,30,31].

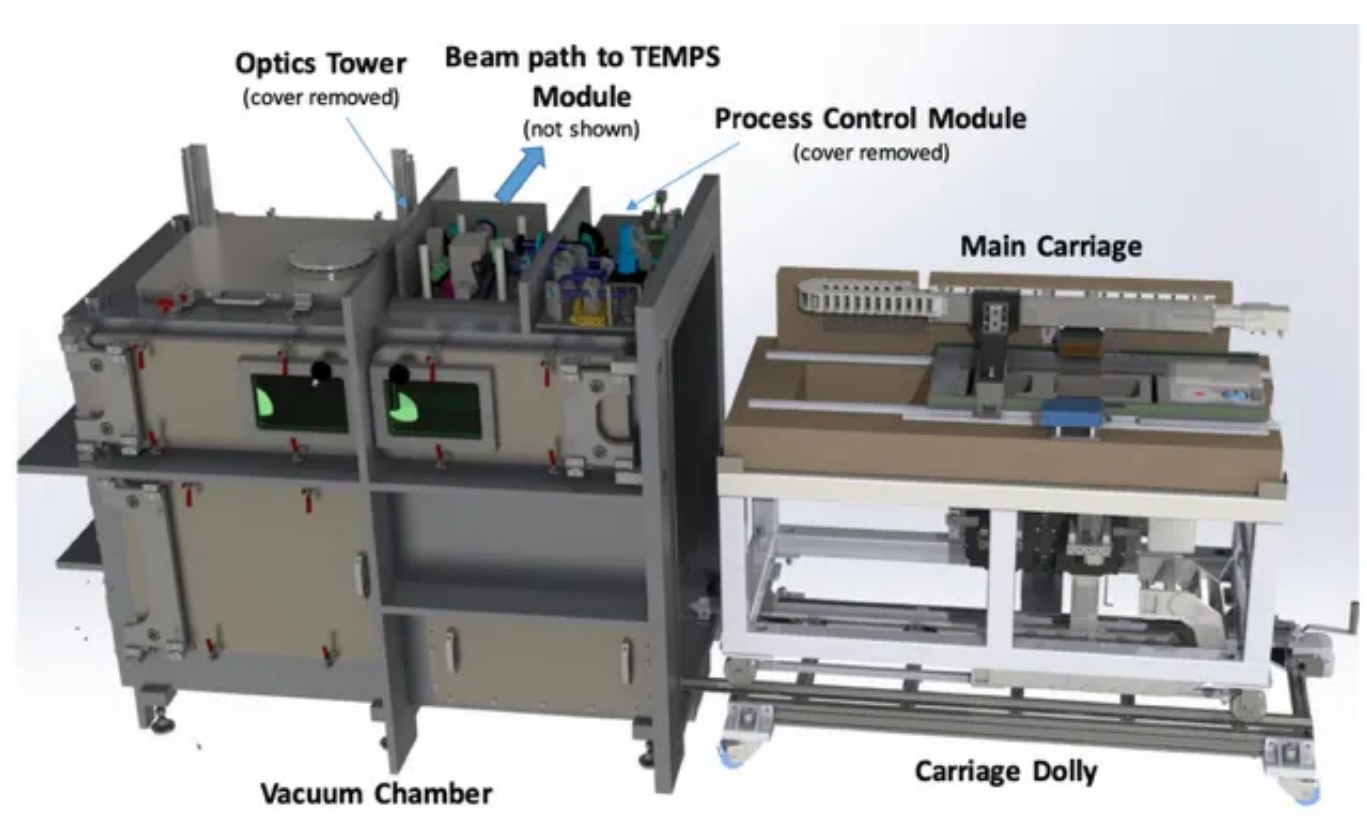

**FIGURE 1:** NIST ADDITIVE MANUFACTURING METROLOGY TESTBED (AMMT) USED FOR MELT POOL IMAGE ACQUISITION

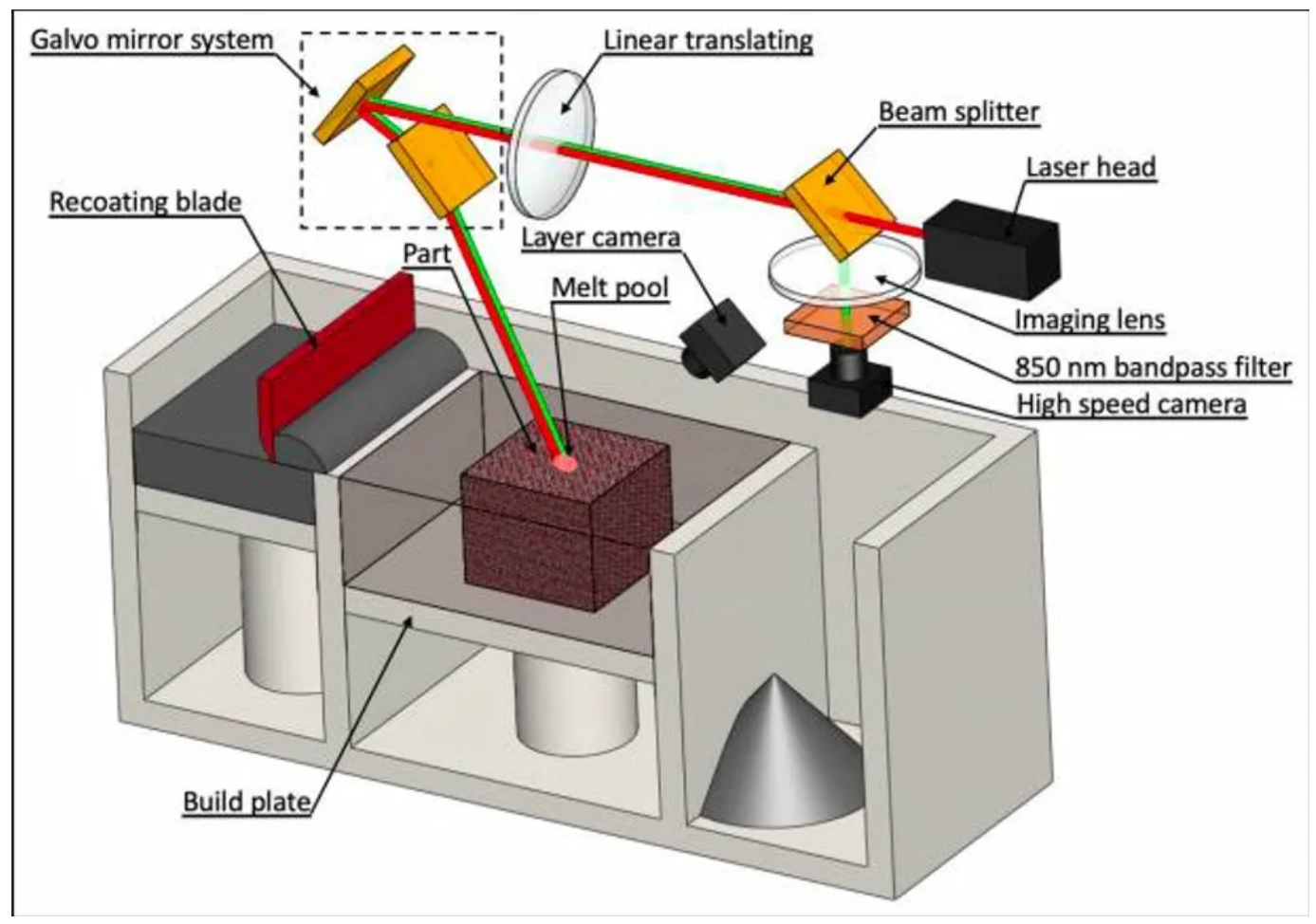

**FIGURE 2:** LPBF BUILD CHAMBER SCHEMATIC SHOWING LASER SCANNING SYSTEM, MELT POOL, AND HIGH-SPEED CAMERA FOR IN-SITU MONITORING

The dataset consists of 1,200 grayscale melt pool images captured during laser powder bed fusion on an instrumented testbed, with each frame cropped to 120 pixels × 120 pixels. Images are evenly split into two classes: 600 labeled normal and 600 labeled abnormal, representing stable and unstable melt pool states, respectively.

Normal and abnormal labels were assigned manually based on expert visual assessment and extensive prior calibration of the laser system. Under constant energy input, a normal melt pool is expected to exhibit a consistent tear-drop shape with stable size and minimal spatter or plume. Images were labeled as abnormal when deviations in melt pool shape, size, aspect ratio, spatter, or plume appearance were observed. It should be noted that these labels are based purely on in-situ image characteristics and domain expertise; no post-process validation (e.g., CT scanning or optical microscopy) was performed to correlate labels with actual porosity or defects.

After removing system-generated metadata files and verifying dimensions, all 1,200 images are retained and organized into class-specific directories and a manifest comma-separated values (CSV) file.

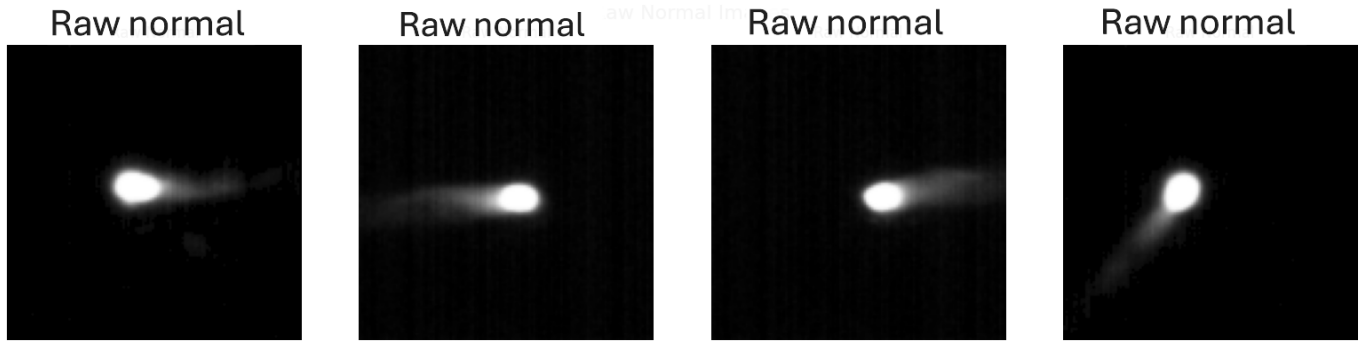

**FIGURE 3:** RAW NORMAL IMAGES

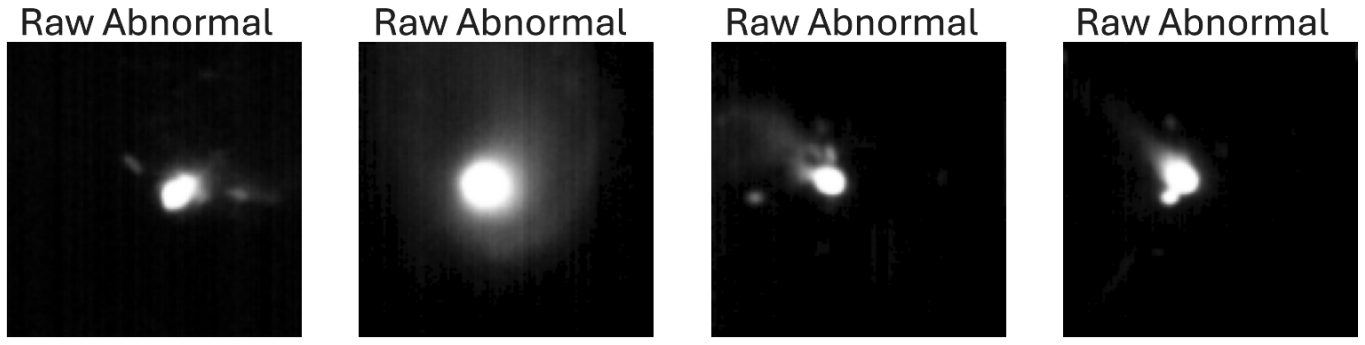

**FIGURE 4:** RAW ABNORMAL IMAGES

### 3.2 Feature representation

Images are resized to 224 pixels × 224 pixels for transfer-learning models, and use pretrained backbones (ResNet50, EfficientNetB0, MobileNetV2) without their original classification heads. Global Average Pooling converts these feature maps to fixed-length vectors, like the 1280-dimensional descriptor from EfficientNetB0 for melt pool images. In the hybrid approach, EfficientNetB0 vectors are used as inputs to a Random Forest classifier, separating feature extraction from the final decision layer. For the baseline Random Forest on raw pixels, images are flattened to 150,528-dimensional vectors without feature extraction.

### 3.3 Data preprocessing strategy

All images were first validated and then split using stratified 80/20 train-test splitting (960/240), preserving 50/50 class balance in each subset. The training portion was further divided into a 90/10 train-validation split. During loading, images were resized to 224 pixels × 224 pixels, batched into groups of 64, shuffled with a fixed random seed, and processed through an optimized data pipeline. Label-preserving data augmentation—random horizontal flips, rotations, zooms, brightness, and contrast adjustments—was applied only to training data to simulate process variability while leaving validation and test sets unchanged.

### 3.4 Selected machine learning models

ResNet50, EfficientNetB0, and MobileNetV2 predictive head:

$$y' = \sigma(w^{\top} h + b) \tag{1}$$

Parameters:

***h***: Global-average-pooled feature vector from the CNN backbone

***w***: Learned weight vector (one weight per feature in $h$)
***b***: Learned bias term
***y'***: Predicted probability of abnormal melt pool (0 to 1)
***σ***: Sigmoid activation function

Random Forest predictive head:

$$\hat{P}(y = 1|h) = \frac{1}{T}\sum_{t=1}^{T} \hat{P}_t(y = 1|h) \tag{2}$$

$$\hat{y} = \begin{cases} 1\ (abnormal)\ if\ \hat{P}(y = 1|h) > \tau \\ 0\ (normal)\ otherwise \end{cases} \tag{3}$$

Parameters:

***h***: 1280-dimensional feature vector from EfficientNetB0 + Global Average Pooling

***T***: Number of decision trees (300–500)
$\hat{P}_t(y = 1|h)$: Probability of abnormal class predicted by tree $t$
$\hat{P}(y = 1|h)$: Forest-averaged probability of abnormal class (0 to 1)
$\tau$: is the decision threshold, which we set to $\tau = 0.58$

### 3.5 Training strategy

For ResNet50, EfficientNetB0, and MobileNetV2, the convolutional base is frozen, and only the newly added classification head (Global Average Pooling + dropout + sigmoid dense layer) is trained, implementing a lightweight transfer-learning strategy appropriate for small datasets. All deep models are trained for up to 20 epochs with early stopping (patience 6) based on validation accuracy to prevent overfitting and unnecessary compute.

For the hybrid Random Forest model, EfficientNetB0 is used purely as a feature extractor; features are computed once for the train and test sets, then a Random Forest with 500 trees is trained on the resulting 1280-dimensional vectors. The baseline Raw Random Forest is trained directly on flattened 150,528-dimensional pixel vectors with 300 trees. Model performance is evaluated on the held-out test set using accuracy, precision, recall, F1-score, and ROC-AUC, alongside training time, per-image inference latency, and memory usage to capture both predictive quality and deployment feasibility.

### 3.6 Rationale behind model selection

The five evaluated architectures span a range of complexity and computational cost. ResNet50 (25 M parameters) uses residual connections to address vanishing gradients in deep networks, typically achieving (90 % to 95 %) accuracy on binary tasks but with higher computational overhead [6]. EfficientNetB0 (5 M parameters) employs compound scaling to balance depth, width, and resolution, delivering (92 % to 97 %) accuracy with superior parameter efficiency suitable for resource-constrained applications [3,8]. MobileNetV2 prioritizes inference speed through depth wise convolutions and inverted residual blocks with 3.5 M parameters, achieving (85 % to 93 %) accuracy optimized for mobile and edge deployment [13]. All three leverage pre-trained ImageNet weights to improve performance on limited training data [6,8].

The hybrid Random Forest approach decouples feature extraction from classification by using frozen EfficientNetB0 to generate 1280-dimensional vectors, which a Random Forest ensemble then classifies [9]. This exploits deep networks' hierarchical feature learning while benefiting from Random Forests' sample efficiency, fast CPU-only inference, and interpretability, achieving (90 % to 98 %) accuracy on binary tasks with high-quality features [9]. The baseline Raw Random Forest, trained directly on flattened pixel vectors, isolates the contribution of deep features versus traditional ensemble methods. Alternative architectures (VGG16, InceptionV3, DenseNet, Vision Transformers, SVM, XGBoost) were excluded: deeper CNNs overfit small datasets without accuracy gains, attention models require massive pre-training data, and classical methods like SVM/XGBoost show slower training on high-dimensional features without demonstrable advantages [8,9].

## 4. EXPERIMENTS

### 4.1 Experimental Setup

Experiments were conducted in Google Colab with free-tier GPU access (typically T4 or K80) using TensorFlow 2.16 and scikit-learn 1.3. All models were trained with a batch size of 64 and up to 20 epochs with early stopping (patience = 6). The unified evaluation protocol measured classification performance alongside training time (minutes), inference latency (ms/image), and approximate GPU/CPU memory usage to assess deployment feasibility.

### 4.2 Data Splitting Strategy

The 1,200-image dataset was first divided into training and test sets using a stratified 80/20 split, preserving the 50/50 balance between normal and abnormal images in both partitions. This produced 960 training images (480 normal, 480 abnormal) and 240 test images (120 normal, 120 abnormal). From the 960 training images, 10 % were then set aside as a validation subset, resulting in 864 images for model training and 96 images for validation, while the test set was kept completely untouched until the final evaluation.

### 4.3 Exploratory Data Analysis (EDA)

EDA confirmed dataset quality: all 960 training images are exactly 120 pixels × 120 pixels with perfect class balance and no corrupted files after cleanup of operating system (OS) artifacts. Width/height distributions show zero variance (std = 0), and aspect ratios are uniformly 1.0. Visual inspection and per-class average images reveal subtle but consistent shape/texture differences between normal (stable, symmetric pools) and abnormal (irregular, elongated) melt pools.

### 4.4 Model Benchmarking Procedure

Phase 1 (validation benchmarking): All five models were trained on the 864/96 train–validation split to identify the most promising candidates. The datasets were processed through an optimized TensorFlow data pipeline that cached batches and prefetched them to keep the GPU fed, and each transfer-learning model applied its corresponding preprocessing function for ResNet, EfficientNet, or MobileNet before inference. The hybrid Random Forest model used 300 trees during this validation phase, while the Raw Random Forest baseline was trained on flattened pixel vectors.

Phase 2 (Final evaluation): Top models were retrained on the full 960-image training set (with augmentation) and evaluated once on the held-out 240-image test set. The hybrid Random Forest used EfficientNetB0 features extracted from the full training set with 500 trees for the final model, while the Raw Random Forest retained 300 trees on raw pixel features.

### 4.5 Performance Metrics

Models were evaluated using standard binary classification metrics: accuracy, precision, recall, F1-score, and ROC-AUC. Resource metrics included training time, per-image inference time, and peak GPU/CPU memory usage. Inference time was measured by recording the total time to predict on the validation set (96 images) and test set (240 images) and dividing by the number of images, capturing end-to-end model execution latency excluding data loading overhead.

**Accuracy**: Fraction of correct predictions across all samples

$$Acc = \frac{TP + TN}{TP + TN + FP + FN} \tag{4}$$

**Precision**: Fraction of predicted abnormal cases that are actually abnormal (minimizes false alarms)

$$Prec = \frac{TP}{TP + FP} \tag{5}$$

**Recall**: Fraction of actual abnormal cases correctly identified (ensures defect detection)

$$Rec = \frac{TP}{TP + FN} \tag{6}$$

**F1-Score**: Harmonic mean of precision and recall; our primary metric balancing both concerns

$$F1 = 2 \times \frac{Prec \times Rec}{Prec + Rec} \tag{7}$$

**AUC-ROC**: Area under receiver operating characteristic curve, i.e. the model's ability to rank abnormal vs normal across all thresholds

$$AUC = \int_0^1 TPR(\tau)\, dFPR(\tau) \tag{8}$$

Where:

$\tau$ is the classification threshold

TPR($\tau$) = $\frac{TP(\tau)}{TP(\tau)+FN(\tau)}$ (True positive rate/recall)

FPR($\tau$) = $\frac{FP(\tau)}{FP(\tau)+TN(\tau)}$ (False positive rate)

$$AUC \approx \sum_{i=1}^{n-1} \left(\frac{FPR_{i+1} - FPR_i}{2}\right) \times (TPR_i + TPR_{i+1})$$
(9)

Where:
**TP (True Positives)**: Abnormal melt pools correctly classified as abnormal
**TN (True Negatives)**: Normal melt pools correctly classified as normal
**FP (False Positives)**: Normal melt pools incorrectly classified as abnormal
**FN (False Negatives)**: Abnormal melt pools incorrectly classified as normal
**TPR (True Positive Rate / Recall)**: TP/(TP + FN)
**FPR (False Positive Rate)**: FP/(FP + TN)

# 5. RESULTS

## 5.1 Overall model performance

Across all experiments, the deep feature-plus-Random Forest hybrid clearly outperformed the transfer learning baselines on the held-out test set. On the 960-image training set, all deep models eventually reached moderate to high training accuracy, but their generalization to validation and test data differed substantially: MobileNetV2 tended to overfit or converge to unstable decision boundaries, while ResNet50 and EfficientNetB0 showed steadier learning but still lagged behind the hybrid approach in F1 and AUC.

**Table 1:** Model validation results

| **Model** | **F1** | **AUC** | **Accuracy** | **Precision** | **Recall** | **Inference (ms/img)** |
|---|---|---|---|---|---|---|
| RF + EN | **0.940** | **0.994** | **0.948** | **0.951** | **0.929** | **0.68** |
| ENB0 | 0.857 | 0.948 | 0.875 | 0.857 | 0.857 | 65.11 |
| Raw RF | 0.800 | 0.947 | 0.844 | 0.909 | 0.714 | 0.95 |
| RN50 | 0.617 | 0.575 | 0.521 | 0.474 | 0.881 | 152.95 |
| MNV2 | 0.508 | 0.547 | 0.656 | 0.680 | 0.405 | 32.13 |

On the 240-image test set, the hybrid Random Forest model built on EfficientNetB0 features achieved the strongest overall performance (F1 = 0.9451, accuracy = 0.9458, AUC = 0.9904), correctly classifying almost all melt pools and clearly separating normal from abnormal cases. ResNet50 with transfer learning was the best among the purely deep models (F1 = 0.7500, accuracy = 0.6750), but it required substantially longer training and inference times (179.51 ms/image vs. 1.15 ms/image for the hybrid). EfficientNetB0 with transfer learning alone also performed well (F1 = 0.7958, accuracy = 0.7583) and was particularly strong at catching abnormal cases, although it produced more false alarms. MobileNetV2 achieved only moderate performance (F1 = 0.6667, accuracy = 0.5000) and tended to behave as if the data were imbalanced, even though the classes were perfectly balanced. The Raw Random Forest baseline trained on pixel features achieved reasonable performance (F1 = 0.7027, accuracy = 0.7250) but could not match the hybrid approach.

**Table 2:** Model test results

| **Model** | **F1** | **AUC** | **Accuracy** | **Precision** | **Recall** | **Inference (ms/img)** |
|---|---|---|---|---|---|---|
| RF + EN | **0.945** | **0.990** | **0.946** | **0.957** | 0.933 | 1.15 |
| ENB0 | 0.796 | 0.939 | 0.758 | 0.689 | 0.942 | 85.47 |
| RN50 | 0.750 | 0.935 | 0.675 | 0.609 | 0.975 | 179.51 |
| Raw RF | 0.703 | 0.828 | 0.725 | 0.765 | 0.650 | **0.99** |
| MNV 2 | 0.667 | 0.418 | 0.500 | 0.500 | **1.00** | 85.43 |

## 5.2 Confusion matrices and error patterns

On the validation split, confusion matrices revealed systematic differences between the architectures. ResNet50 tended either to over-predict the abnormal class or misclassify many true abnormal cases as normal, leading to poor balance between precision and recall. MobileNetV2, despite reasonable accuracy in some runs, exhibited highly skewed predictions with many false negatives or false positives, consistent with its unstable F1 and AUC scores.

EfficientNetB0 transfer learning improved both the total number of correct predictions and the balance between the two classes, but still produced a noticeable number of false positives due to its tendency toward very high recall. The hybrid Random Forest on EfficientNet features showed the most compact confusion matrix, with both high true-normal and high true-abnormal counts and relatively few misclassifications in either quadrant, on both validation and test sets. On the test set, the hybrid model misclassified only (13 out of 240) images (5 false positives, 8 false negatives).

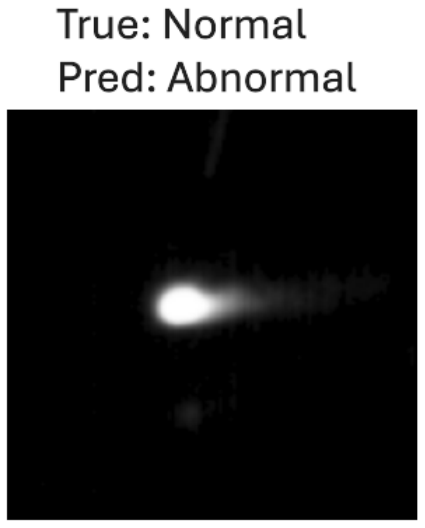

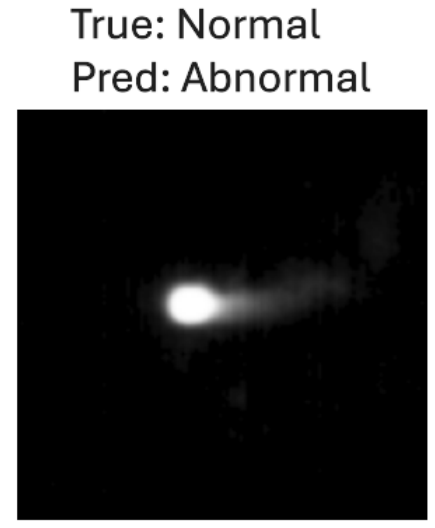

**FIGURE 5:** FALSE POSITIVES - NORMAL IMAGES WRONGLY PREDICTED ABNORMAL BY THE HYBRID MODEL (VALIDATION SPLIT)

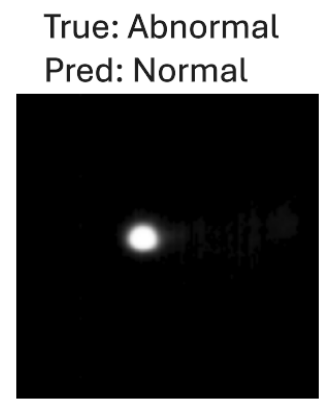

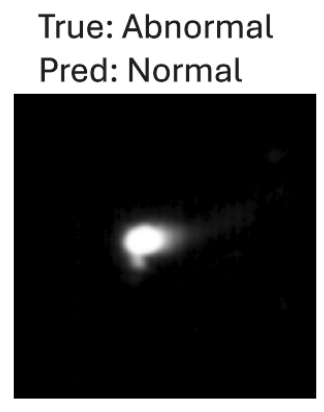

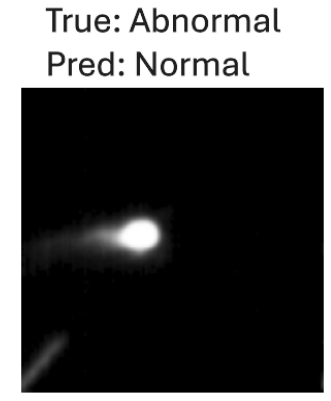

**FIGURE 6:** FALSE NEGATIVES - ABNORMAL IMAGES WRONGLY PREDICTED NORMAL BY THE HYBRID MODEL (VALIDATION SPLIT)

### 5.3 Quantitative results and interpretation

From a resource perspective, the hybrid Random Forest achieved its superior performance with short training times (around 1.6 minutes) and sub-millisecond per-image inference (1.15 ms) on CPU, while not requiring a GPU at deployment. In contrast, ResNet50 and EfficientNetB0 models needed significantly longer training (62.7 minutes and 26.0 minutes, respectively) and had substantially higher inference latency (179.51 ms and 85.47 ms), reflecting their larger parameter counts and sequential layer computations. The Raw Random Forest baseline, while fast (0.99 ms inference), lacked the representational capacity provided by deep features, resulting in lower F1 (0.7027) and AUC (0.8283).

The combination of high F1, near-perfect AUC, and low inference cost demonstrates that pretrained EfficientNet features paired with a Random Forest classifier provide the best trade-off between accuracy, robustness, and deployability for this small, balanced melt pool dataset. This supports the broader observation that hybrid "deep features + classical ML" pipelines can outperform purely end-to-end CNNs in data-limited industrial monitoring scenarios where both precision (avoiding false alarms) and recall (catching true anomalies) are critical.

## 6. SUMMARY

Laser powder bed fusion (LPBF) faces persistent quality challenges due to melt pool anomalies that lead to porosity and defects in metal components. Real-time detection of abnormal melt pools is critical for process control, yet existing deep learning approaches struggle to balance high classification accuracy with the millisecond-level inference speeds required for industrial deployment. This work addresses the question: can a hybrid ML architecture achieve both superior accuracy and computational efficiency for melt pool anomaly detection on limited training data?

We propose a hybrid pipeline that decouples feature extraction from classification by using a frozen EfficientNetB0 backbone to generate 1280-dimensional feature vectors, which are then classified by a Random Forest ensemble. This approach is benchmarked against three end-to-end transfer learning models (ResNet50, EfficientNetB0, MobileNetV2) and a baseline Random Forest trained on raw pixels. All models are evaluated on a balanced dataset of 1,200 melt pool images (originally 120 pixels ×120 pixels) split into 960 training images and 240 test images, with assessment based on accuracy, precision, recall, F1-score, ROC-AUC, training time, and inference latency.

The proposed hybrid model substantially outperforms all alternatives on the held-out test set, achieving F1 = 0.9451, accuracy = 0.9458, and AUC = 0.9904 with only 1.15 ms per-image inference on CPU. In contrast, purely deep learning models exhibit both lower accuracy and dramatically higher inference times: ResNet50 (F1 = 0.7500, 179.51 ms), EfficientNetB0 (F1 = 0.7958, 85.47 ms), and MobileNetV2 (F1 = 0.6667, 85.43 ms). The Raw Random Forest baseline (F1 = 0.7027) confirms that deep features are essential for high performance. These results demonstrate that the hybrid approach provides the optimal balance between accuracy and speed for real-time LPBF monitoring.

Primary limitations of this study are as follows. The dataset contains 1,200 images and comes from a single experimental session on the NIST AMMT platform using Nickel superalloy 625 powder. All images have uniform dimensions (120 × 120 pixels) and were captured under limited process parameter variation with constant energy input. The dataset is balanced (i.e., 50/50 normal/abnormal), which does not reflect real-world LPBF conditions where anomalies are rare events. Furthermore, the normal and abnormal labels were assigned manually based on expert visual assessment and domain experience. A normal melt pool was defined as having a consistent tear-drop shape with stable size and minimal spatter or plume; images deviating in shape, size, aspect ratio, spatter, or plume were labeled abnormal. It should be noted that these labels are based solely on in-situ image characteristics and have not been cross-validated against post-process destructive testing such as CT scans or microscopy to confirm actual porosity or defects.

Secondly, results are based on a single train-test split without k-fold cross-validation, making it difficult to fully assess performance stability and robustness. The current approach evaluates individual static frames and does not incorporate temporal modeling of melt pool sequences, even though anomalies often evolve dynamically over consecutive frames. The study also used frozen transfer-learning backbones without fine-tuning and performed limited hyperparameter optimization due to computational constraints.

Lastly, the systematic benchmarking and explicit focus on deployability (latency and resource usage) are notable strengths of this study. However, the core hybrid architecture combining pretrained CNN features with a Random Forest classifier has been previously explored in domains such as medical imaging and remote sensing. The primary contribution of this work therefore lies in its application and rigorous evaluation for LPBF melt pool monitoring rather than in methodological innovation. Furthermore, the experiments were conducted exclusively on Nickel superalloy 625 using the NIST AMMT research platform. Generalization of these results to other materials, commercial LPBF systems, or different imaging conditions should therefore be approached with caution and validated in future studies.

## DISCLAIMER

Certain commercial systems are identified in this paper. Such identification does not imply recommendation or endorsement by NIST; nor does it imply that the products identified are necessarily the best available for the purpose. Further, any opinions, findings, conclusions, or recommendations expressed in this material are those of the authors and do not necessarily reflect the views of NIST or any other supporting U.S. government or corporate organizations.

## REFERENCES

[1] Rehman, A. U., Pitir, F., & Salamci, M. U. (2021). Full-Field Mapping and Flow Quantification of Melt Pool Dynamics in Laser Powder Bed Fusion of SS316L. Materials, 14(21), 6264. https://doi.org/10.3390/ma14216264

[2] Naderi Yaghouti, A. R., & Shalbaf, A. (2025). Classification of Nonalcoholic Fatty Liver Grades using Pre-Trained Convolutional Neural Networks and a Random Forest Classifier on B-Mode Ultrasound Images. Journal of Biomedical Physics & Engineering, 15(5), 437. https://doi.org/10.31661/jbpe.v0i0.2307-1646

[3] Rahman, M. S., Sattar, N. S., Ahmed, R. U., Ciaccio, J., and Chakravarty, U. K. (August 6, 2024). "A Machine Learning Framework for Melt-Pool Geometry Prediction and Process Parameter Optimization in the Laser Powder-Bed Fusion Process." ASME. J. Eng. Mater. Technol. October 2024; 146(4): 041006. https://doi.org/10.1115/1.4065687

[4] Wang, R., Garcia, D., Kamath, R. R., Dou, C., Ma, X., Shen, B., Choo, H., Fezzaa, K., Yu, H. Z., & Kong, Z. (2022). In situ melt pool measurements for laser powder bed fusion using multi sensing and correlation analysis. Scientific Reports, 12(1), 13716. https://doi.org/10.1038/s41598-022-18096-w

[5] Jiayi Zhang, Ce Yin, Yiyang Xu, Swee Leong Sing. Machine learning applications for quality improvement in laser powder bed fusion: A state-of-the-art review. International Journal of AI for Materials and Design 2024, 1(1), 26–43. https://doi.org/10.36922/ijamd.2301

[6] Xing, W., Chu, X., Lyu, T., Lee, C., Zou, Y., & Rong, Y. (2022). Using convolutional neural networks to classify melt pools in a pulsed selective laser melting process. Journal of Manufacturing Processes, 74, 486-499. https://doi.org/10.1016/j.jmapro.2021.12.030

[7] Z. Yang, Y. Lu, H. Yeung and S. Krishnamurty, "Investigation of Deep Learning for Real-Time Melt Pool Classification in AM," 2019 IEEE 15th International Conference on Automation Science and Engineering (CASE), Vancouver, BC, Canada, 2019, pp. 640-647, doi: 10.1109/COASE.2019.8843291. keywords: {Powders;Monitoring;Real-time systems;Process control;Cameras;Shape;Laser beams},

[8] Zhang, S., Yang, H., Yang, Z., and Lu, Y. (August 6, 2024). "Engineering-Guided Deep Learning of Melt-Pool Dynamics for AM Quality Monitoring." ASME. J. Comput. Inf. Sci. Eng. October 2024; 24(10): 101002. https://doi.org/10.1115/1.4066026

[9] Ribeiro, K.S.B., Núñez, H.H.L., Venter, G.S. et al. A hybrid machine learning model for in-process estimation of printing distance in laser Directed Energy Deposition. Int J Adv Manuf Technol 127, 3183–3194 (2023). https://doi.org/10.1007/s00170-023-11582-z

[10] Yuan, Bodi, Guss, Gabriel M., Wilson, Aaron C., Hau-Riege, Stefan P., DePond, Phillip J., McMains, Sara, Matthews, Manyalibo J., & Giera, Brian (2018). Machine-Learning-Based Monitoring of Laser Powder Bed Fusion. Advanced Materials Technologies, 3(12). https://doi.org/10.1002/admt.201800136

[11] Yang, J., Huang, K., & Lin, P. (2023). Three-Dimensional Printing Quality Inspection Based on Transfer Learning with Convolutional Neural Networks. Sensors (Basel, Switzerland), 23(1), 491. https://doi.org/10.3390/s23010491

[12] In-situ melt pool characterization via thermal imaging for defect detection in directed energy deposition. (2023). arXiv Preprint. https://arxiv.org/html/2411.12028v2

[13] Kuriachen, B., Jeyaraj, R., Raphael, D., Ashok, P., Sundari, P. S., & Paul, A. (2025). Defect detection in fused deposition modelling using lightweight convolutional neural networks. Engineering Applications of Artificial Intelligence, 141, 109802. https://doi.org/10.1016/j.engappai.2024.109802

[14] Barrutia, A., Elzaurdi, M., & San Sebastian, M. (2024). Melt pool monitoring and machine learning approaches for anomaly detection in PBF-LB. Procedia CIRP, 124, 785-788. https://doi.org/10.1016/j.procir.2024.08.225

[15] A data-efficient sequential learning framework for melt pool defect classification in laser powder bed fusion. (2024). arXiv Preprint. https://arxiv.org/abs/2411.10822

[16] Yaqoob, M. K., Ali, S. F., Bilal, M., Hanif, M. S., & Al-Saggaf, U. M. (2021). ResNet Based Deep Features and

Random Forest Classifier for Diabetic Retinopathy Detection. Sensors, 21(11), 3883. https://doi.org/10.3390/s21113883

[17] Muñoz-Saavedra, L., Escobar-Linero, E., Civit-Masot, J., Luna-Perejón, F., Civit, A., & Domínguez-Morales, M. (2023). A Robust Ensemble of Convolutional Neural Networks for the Detection of Monkeypox Disease from Skin Images. Sensors, 23(16), 7134. https://doi.org/10.3390/s23167134

[18] Ma, X., Shan, J., Ning, F., Li, W., & Li, H. (2023). EFFNet: A skin cancer classification model based on feature fusion and random forests. PLOS ONE, 18(10), e0293266. https://doi.org/10.1371/journal.pone.0293266

[19] Xiao, Y., Wang, X., Yang, W., Yao, X., Yang, Z., Lu, Y., Wang, Z., & Chen, L. (2024). Data-driven prediction of future melt pool from built parts during metal AM. Additive Manufacturing, 93, 104438. https://doi.org/10.1016/j.addma.2024.104438

[20] Satish Chandra, N., Raj, S., & Mahanand, B. S. (2025). NGBoost Classifier Using Deep Features for Pneumonia Chest X-Ray Classification. Applied Sciences, 15(17), 9821. https://doi.org/10.3390/app15179821

[21] Harbig, J., Wenzler, D. L., Baehr, S., Kick, M. K., Merschroth, H., Wimmer, A., Weigold, M., & Zaeh, M. F. (2022). Methodology to Determine Melt Pool Anomalies in Powder Bed Fusion of Metals Using a Laser Beam by Means of Process Monitoring and Sensor Data Fusion. Materials, 15(3), 1265. https://doi.org/10.3390/ma15031265

[22] Settu, P., & Ramaiah, M. (2025). A data driven comparison of hybrid machine learning techniques for soil moisture modeling using remote sensing imagery. Scientific Reports, 15, 43170. https://doi.org/10.1038/s41598-025-27225-0

[23] Tang, J., Tan, K., Tang, J., Zhao, Z., Zhang, X., & Chen, X. (2025). Uncertainty-driven trustworthy identification paradigm for unstable melt pool state based on acoustic emission in LPBF. Additive Manufacturing, 109, 104887. https://doi.org/10.1016/j.addma.2025.104887

[24] An in situ pyrometry and ex situ X-ray radiography correlation study of pore formation in laser powder bed fusion. (2019). Journal of Manufacturing Processes, 45, 1–12. https://www.osti.gov/servlets/purl/1635089

[25] Prasad Vallabh, C. K., & Zhao, X. (2023). Continuous Comprehensive Monitoring of Melt Pool Morphology Under Realistic Printing Scenarios with Laser Powder Bed Fusion. 3D Printing and AM, 10(1), 101. https://doi.org/10.1089/3dp.2021.0060

[26] Shaikh, M. A., Zaman, H., & Asif, A. (2025). Comparative study of CNN architectures for binary classification of horses and motorcycles in the VOC 2008 dataset (arXiv Preprint No. 2511.04344). arXiv. https://arxiv.org/abs/2511.04344

[27] Lane, Brandon, Mekhontsev, Sergey, Grantham, Steven, Vlasea, ML, Whiting, Justin, Yeung, Ho, Fox, Jason, Zarobila, Clarence, Neira, Jorge, McGlauflin, Michael et al. “Design, developments, and results from the NIST AM metrology testbed (AMMT).” (2016).

[28] Yeung, Ho, Lane, Brandon M, Donmez, MA, Fox, Jason C and Neira, Jorge. “Implementation of advanced laser control strategies for powder bed fusion systems.” Procedia Manufacturing Vol. 26 (2018): pp. 871–879.

[29] Yeung, Ho, Hutchinson, Keely and Lin, Dong. “Design and implementation of laser powder bed fusion AM testbed control software.” (2021).

[30] Deisenroth, David C, Neira, Jorge, Weaver, Jordan and Yeung, Ho. “Effects of shield gas flow on meltpool variability and signature in scanned laser melting.” International Manufacturing Science and Engineering Conference, Vol. 84256: p. V001T01A017. 2020. American Society of Mechanical Engineers.

[31] Lane, Brandon, Lane, Brandon, Moylan, Shawn, Yeung, Ho, Neira, Jorge and Chavez-Chao, Josephine. Quasi-static position calibration of the galvanometer scanner on the AM metrology testbed. US Department of Commerce, National Institute of Standards and Technology (2020)